\documentclass[10pt,twocolumn,letterpaper]{article}

\usepackage{iccv}
\usepackage{times}
\usepackage{epsfig}
\usepackage{graphicx}
\usepackage{amsmath}
\usepackage{amssymb}
\usepackage{authblk}
\usepackage{blindtext}
\def\Plus{\texttt{+}}
\def\Minus{\texttt{-}}

\usepackage[breaklinks=true,bookmarks=false]{hyperref}

\iccvfinalcopy 

\def\iccvPaperID{****} 

\ificcvfinal\fi

\begin{document}

\title{Designovel's system description for Fashion-IQ challenge 2019}

\author[1,2]{\textsuperscript{$\dagger$}Jianri Li}
\author[1]{\textsuperscript{$\ddag$}Jae-whan Lee}
\author[1,3]{\textsuperscript{$\ddag$}Woo-sang Song}
\author[1,3]{\textsuperscript{$\ddag$}Ki-young Shin}
\author[2]{\textsuperscript{$\dagger$}Byung-hyun Go}

\affil[1]{Desiginovel}
\affil[2]{Department of Computer Science and Engineering, Pohang University of Science and Technology}
\affil[3]{Department of Creative IT Engineering, Pohang University of Science and Technology}
\affil[$\dagger$]{\textit {\{skywalker,briango28\}@postech.ac.kr}}
\affil[$\ddag$]{\textit {\{whynhle,woosang,shinky\}@designovel.com}}
\renewcommand\Authands{ and }

\maketitle
\ificcvfinal\thispagestyle{empty}\fi

\begin{abstract}
This paper describes Designovel's systems submitted to the Fashion IQ 
Challenge 2019. The goal of the challenge is to build an image retrieval system
in which the input query is a candidate image with two text phrases describing users' feedback about
visual differences between the candidate image and the search target. We built the systems by combining methods
from recent work on deep metric learning, multi-modal retrieval and natural
language processing. First, we encode both candidate and target images with 
CNNs into high-level representations, and encode text descriptions
to a single text vector using a Transformer-based encoder. Then we compose
the candidate image vector and text representation into a single vector which is expected to
be biased toward the target image vector. Finally, we compute cosine similarities 
between the composed vector and encoded vectors of the whole dataset, and rank
them in descending order to get a ranked list. We experimented with the Fashion IQ 2019
dataset with various hyperparameters, achieving a 39.12\% average recall 
with a single model and 43.67\% average recall with an ensemble of 16 models on the test dataset. 
\end{abstract}

\section{Introduction}
We participated in the Fashion IQ Challenge 2019 by building image+text to image retrieval
systems on fashion items in three pre-defined categories: dress, shirt and toptee. 

\begin{figure}[t]
\begin{center}
\includegraphics[width=1.0\linewidth]{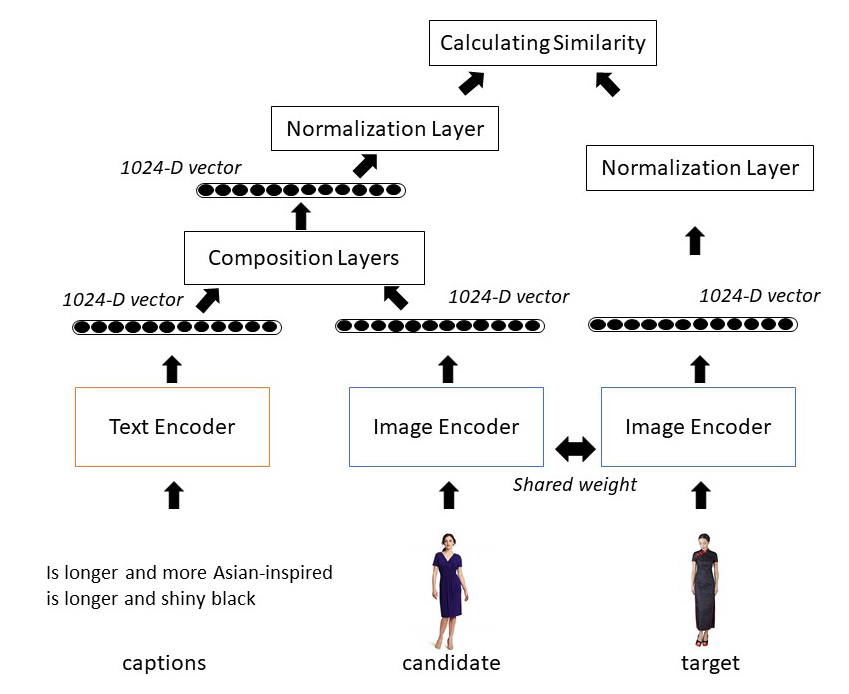}
\end{center}
   \caption{Overview of Designovel's system.}
\label{fig:fig-1}
\end{figure}

Our baseline system consists of an image encoder, a text encoder and a composition module as shown
in Figure~\ref{fig:fig-1}. The image encoder is based on the VGG network \cite{vgg} with landmark-driven 
attention layers \cite{Liu}, and we use BERT \cite{bert} as the text encoder. The composition module 
is based on the TIRG method introduced in \cite{Nam}.

We use the Fashion IQ dataset \cite{iq} for training our system on this task.
As additional data, we use the Deepfashion dataset \cite{deepfashion} to 
pre-train the image encoder and use an in-house fashion-domain corpus to pre-train the text encoder.

We trained our systems using data augmentation techniques, regularizations like dropout and label smoothing.

We evaluated our systems by ensembling several models trained separately. As the task suggested, 
the evaluation process measures recall percentage for each of the top 10 and top 50 of ranked results for three categories: dress, shirt and toptee.

In the test phase, our system achieved 43.67\% average recall and ranked the third place among participants.

In following sections, we will describe details of our method, experiment settings, evaluation results and conclusion.

\section{Method}
Our baseline model includes three main parts: the image encoder, the text encoder and the composition layers. \par
For each candidate and target image pair, we proceed with the following steps: \par
1) Feed both candidate and target image into image encoder and linearly project outputs into the same space, which has 1024 dimensions.

2) Convert caption tokens using pre-trained text embeddings to 384-dimensional vectors and feed them to the text encoder and use the hidden 
representation of the token ``CLS'' as the representation of the whole document.

3) Compose candidate vectors and caption representation into the same space as the image vector's using TIRG composition. By doing this,
we expect the candidate image vector to be biased toward the target image vector so that cosine similarity between candidate and target image vectors is larger than that between candidate and other images which are not the target image.

4) Normalize composed vector and target image vectors with second-order norm.

5) Compute cosine similarities between composed vector and target image vector for final ranking.

In the following subsections, we describe the details of our sub-modules.

\subsection{Image encoder}
The image encoder, a VGG-based convolutional network, is identical to the model introduced in \cite{Liu} except for two differences: 
1) we add batch normalization layers to VGG, 
2) in the attribute prediction layers for pre-training, instead of directly predicting 1000 attributes from a single feature vector, we use Attribute Prediction Network as described in \cite{iq} to predict attributes separately from five feature vectors each corresponding to an attribute category. \par
Before applying the image encoder to the main task, we pre-trained the image encoder with the Deepfashion dataset \cite{deepfashion} with training 
objectives consisting of attribute prediction, category prediction and landmark prediction. We optimized attribute prediction using binary cross 
entropy loss, category prediction with negative log-likelihood loss and landmark prediction with mean squared error loss. We applied 
weights of 20, 1, 10 for each loss respectively and summed up these weighted losses for single-step optimization.

\subsection{Text encoder}
The text encoder, a Transformer-based network, is similar to BERT introduced in \cite{bert} except for three parts:
1) we reduced the size of the model due to the fact that in our task, average text length is shorter and text structure is simpler than those in the original 
paper, 
2) we distinguish the role of each layer in the encoder by restricting the attention range of the self-attention mechanism,
3) in addition to masked language modeling, we use item category prediction as a sub-task instead of next sentence prediction used in the 
original paper. \par
Our text encoder has four self-attentional layers, each with 384 hidden dimensions, six self-attentional heads and 1536 intermediate hidden dimensions. We noticed that the input text is a document which consists of several sentences that are not necessarily in fixed order, therefore we assign different roles to each layer in the encoder. The first two layers are defined as sentence layers in which self-attention is only applied among inter-sentence tokens. The last two layers are defined as document layers, in which self-attention is applied to all tokens in the document with the purpose of capturing information from the whole document. The input format is the same as the original paper: sentences are separated by a special token ``SEP'' and the first position is always the class prediction token ``CLS''. We also randomly shuffled sentence order during training for better robustness.
\subsection{Image-text composition layers}
We use TIRG \cite{Nam} to compose candidate image and caption text. In addition to the original TIRG function, we add category embeddings to distinguish composition
behaviour in three fashion categories. The TIRG function used in our system has the following form:
\begin{equation}
    {\phi}_{xt}^{rg} = w_gf_{gate}({\phi}_x, {\phi}_t, {\phi}_c) + w_rf_{res}({\phi}_x, {\phi}_t, {\phi}_c)
\end{equation}
\begin{equation}
    f_{gate}({\phi}_x, {\phi}_t, {\phi}_c) = {\sigma}(W_{g2}*RELU(W_{g1}*[{\phi}_x, {\phi}_t, {\phi}_c])){\odot}{\phi}_x
\end{equation}
\begin{equation}
    f_{res}({\phi}_x, {\phi}_t, {\phi}_c) = W_{r2}*RELU(W_{r1}*[{\phi}_x, {\phi}_t, {\phi}_c])
\end{equation}

Where ${\phi}_x$, ${\phi}_t$, ${\phi}_c$ denote candidate image vector, caption vector and category embeddings vector, respectively.
\subsection{Loss function}
Our loss function for deep metric learning is similar to \cite{rankedlist}, except that we use cosine similarity instead of euclidean distance. We found this method can achieve slightly higher recall rate than N-pair loss \cite{n-pair} used in \cite{Nam}.

\section{Experiment}
\subsection{Data}
We use the Fashion-IQ dataset for the main task. For training, we simply create pseudo training pairs using images that are never used as a target 
in the training set. For each of those images, we construct a pseudo example in which both the candidate and target images are identical to the original image itself. We also gather phrases from captions of the training data indicating equivalence such as ``exactly same'' or ``is the same item'' using some hand-made rules. Among these phrases, two were randomly selected as the caption for each pseudo example.

For pre-training of the image encoder, we use the Deepfashion dataset \cite{deepfashion}.

\begin{table}
\centering
\begin{tabular}{|p{5cm}||p{2cm}|}
    \hline
    \multicolumn{2}{|c|}{Designovel's Fashion Corpus} \\
    \hline
    \verb|#|documents & 1,120,465 \\
    \verb|#|tokens & 58,613,441 \\
    \verb|#|categories & 112 \\
    Maximum document length & 140 \\
    Average document length & 52 \\
    \hline
\end{tabular}
    \caption{Text corpus for text encoder pre-training}
    \label{table:1}
\end{table}

For pre-training of the text encoder, we use our in-house fashion-domain corpus built from crawling online shopping malls. Details and statistics of the corpus is shown in Table \ref{table:1}. Each document in the corpus is a description about a unique fashion item. The description includes information like motivation from the designer or brand, visual details, colors, components, materials and stitching methods.
\subsection{Data pre-processing and augmentation}
For all images used in our experiments, we use the MMDetection tool \cite{mm} to calculate the bounding box of fashion objects appearing in 
an image and crop the image using these boundaries. The object detector is also trained with the Deepfashion dataset \cite{deepfashion}.

For augmentation, we used random horizontal flips, random angle affine transformations, random horizontal and vertical translations, random distortion and 
random erasing. We found data augmentation could significantly improve performance.
\subsection{Hyperparameters and learning curriculum}
We used Adam \cite{adam} as the optimizer and set the initial learning rate to 5e-5 for composition layers and 5e-6 for image and text encoders. 

We separate the training data according to three fashion categories and trained the model in the order of ``dress, shirt, toptee''. In this case, we define one epoch as an iteration over all three categories.

\subsection{Result}
We experimented on the Fashion-IQ dataset with various settings as shown in Table \ref{table:2}. 
We began with the baseline in which we set the same learning rate 5e-5 on all modules, and gradually adjusted the settings by lowering the learning rate of text and image encoder to 5e-6 (as suggested in \cite{Nam}), applying data augmentation on the training data and ensembling several trained models from different runs. For model ensembling, we first calculated similarity scores with each sub-model separately and then simply averaged these scores.

As final result, we achieved an average recall of 39.12\% with a single model and 43.67\% with an ensemble of 16 models on the test dataset.

\begin{table}
\centering
    \begin{tabular}{|p{3.5cm}||p{1.5cm}|p{1.5cm}|}
        \hline
        \Minus & Validation & Test \\
        \hline
        Baseline & 34.09 & \Minus \\
        \Plus Small LR on encoder & 37.28 & 36.49 \\
        \Plus Data augmentation & 40.84 & 39.14 \\
        \Plus Ensemble (8 models) & 45.00 & 43.52 \\
        \Plus Ensemble (16 models) & 45.86 & 43.67 \\
        \hline
    \end{tabular}
    \caption{Evaluation results (average recall, \%) on Fashion-IQ dataset}
    \label{table:2}
\end{table}
\section{Conclusion}
We participated in the Fashion IQ Challenge 2019 by building an image+text to image retrieval system using methods from recent works and achieved a 43.67\% average recall with an ensemble of 16 models in the test phase. 

By experimenting on various settings, we found that simple data augmentation and model ensembles could significantly improve recall percentage.

As future work, we will focus on the positive example mining method since each candidate can have multiple matched targets, while the given training and validation datasets only indicate a single target which may potentially lead to overfitting. We will also try various ensembling methods instead of simply averaging scores.

{\small
\bibliographystyle{ieee_fullname}
\bibliography{egbib}
}
\end{document}